\pdfoutput=1

\documentclass[11pt]{article}

\usepackage[final]{acl}

\usepackage{times}
\usepackage{latexsym}

\usepackage{minitoc}
\usepackage{titletoc}

\usepackage[T1]{fontenc}

\usepackage[utf8]{inputenc}

\usepackage{microtype}

\usepackage{inconsolata}

\usepackage{bm}
\usepackage{tikzducks}
\usepackage{graphicx}
\usepackage[most]{tcolorbox}
\usepackage{colortbl}
\usepackage{pifont}
\usepackage{enumitem}
\definecolor{Gray}{gray}{0.9}

\usepackage{subfigure}
\usepackage{tikz}

\usepackage{listings}
\definecolor{codeblue}{rgb}{0.25,0.5,0.5}
\definecolor{codekw}{rgb}{0.85, 0.18, 0.50}
\definecolor{tear}{rgb}{25, 25, 112}

\lstdefinestyle{mystyle}{
    backgroundcolor=\color{white},
    basicstyle=\fontsize{7.5pt}{7.5pt}\ttfamily\selectfont,
    columns=fullflexible,
    breaklines=true,
    captionpos=b,
    commentstyle=\fontsize{7.5pt}{7.5pt}\color{codeblue},
    keywordstyle=\fontsize{7.5pt}{7.5pt}\color{codekw},
}
\usepackage{graphicx}
\usepackage{amsmath, amssymb, bm}
\usepackage[ruled,vlined]{algorithm2e} 
\usepackage{algpseudocode}
\usepackage{multirow}
\usepackage{graphicx}
\usepackage{booktabs}
\usepackage{multicol}
\usepackage{hyperref}
\usepackage{amsmath}
\usepackage{caption}
\usepackage{enumitem}
\usepackage{wrapfig}
\usepackage{listings}
\usepackage{pifont}
\usepackage{array}
\usepackage{placeins}
\usepackage{float}
\definecolor{codegreen}{rgb}{0.0, 0.411, 0.243}
\definecolor{codered}{rgb}{0.89, 0.26, 0.20}
\usepackage{hyperref}
\definecolor{dartgreen}{HTML}{00693e}
\definecolor{refcolor}{HTML}{9F363A}

\hypersetup{
    colorlinks=true,
    linkcolor=refcolor,
    citecolor=refcolor,
    filecolor=magenta,      
    urlcolor=refcolor,
    }

\title{ProtoVQA: An Adaptable Prototypical Framework for Explainable Fine-Grained Visual Question Answering}
\author{
 \textbf{Xingjian Diao$^\diamond$$^\clubsuit$},
 \textbf{Weiyi Wu$^\diamond$$^\clubsuit$},
 \textbf{Keyi Kong$^\bigstar$},
 \textbf{Peijun Qing$^\clubsuit$},
 \textbf{Xinwen Xu$^\spadesuit$},
 \\
 \textbf{Ming Cheng\textsuperscript{$^\clubsuit$}}\textbf{,}
 \textbf{Soroush Vosoughi\textsuperscript{$^\clubsuit$}}\textbf{,}
 \textbf{Jiang Gui\textsuperscript{$^\clubsuit$}}
 \\
 $^\clubsuit$Dartmouth College 
 $^\bigstar$Shandong University 
 $^\spadesuit$Harvard University
 \\
   \texttt{\{xingjian.diao, weiyi.wu\}.gr@dartmouth.edu}
}
\begin{document}
\maketitle

\newcommand\blfootnote[1]{%
  \begingroup
  \renewcommand\thefootnote{}\footnote{#1}%
  \addtocounter{footnote}{-1}%
  \endgroup
}

\blfootnote{$^\diamondsuit$contributed equally}

\begin{abstract}
Visual Question Answering (VQA) is increasingly used in diverse applications ranging from general visual reasoning to safety-critical domains such as medical imaging and autonomous systems, where models must provide not only accurate answers but also explanations that humans can easily understand and verify. Prototype-based modeling has shown promise for interpretability by grounding predictions in semantically meaningful regions for purely visual reasoning tasks, yet remains underexplored in the context of VQA. We present \texttt{ProtoVQA}, a unified prototypical framework that (i) learns question-aware prototypes that serve as reasoning anchors, connecting answers to discriminative image regions, (ii) applies spatially constrained matching to ensure that the selected evidence is coherent and semantically relevant, and (iii) supports both answering and grounding tasks through a shared prototype backbone. To assess explanation quality, we propose the Visual–Linguistic Alignment Score (VLAS), which measures how well the model’s attended regions align with ground-truth evidence. Experiments on Visual7W show that \texttt{ProtoVQA} yields faithful, fine-grained explanations while maintaining competitive accuracy, advancing the development of transparent and trustworthy VQA systems.
\end{abstract}

\section{Introduction}
\label{sec:intro}
Visual Question Answering (VQA) is a key challenge in AI, requiring systems to understand and reason about both visual content and natural language queries \cite{zhu2016visual7w, Kafle_2017_ICCV,li2024configure}. Recent advances in vision transformers \cite{dosovitskiy2021an, touvron2021training} have significantly improved performance by enhancing multimodal feature learning, leading to better accuracy on VQA benchmarks.

As VQA systems are applied in critical fields such as medical diagnosis \cite{wang2022knowledge, donnelly2024asymmirai, yang2024neuron}, autonomous driving \cite{ramos2017detecting} and criminal justice \cite{berk2019machine}, model transparency is essential. Current state-of-the-art models operate as black boxes, making it difficult to interpret their reasoning or verify reliability. 
Traditional VQA interpretability approaches, primarily using attention visualization or post-hoc explanation methods, often fail to faithfully represent the internal decision-making process of the model \cite{chen2019looks,ma2023looks,ma2024interpretable}.

\begin{figure*}[tb]
\begin{center}
\includegraphics[width=1\linewidth]{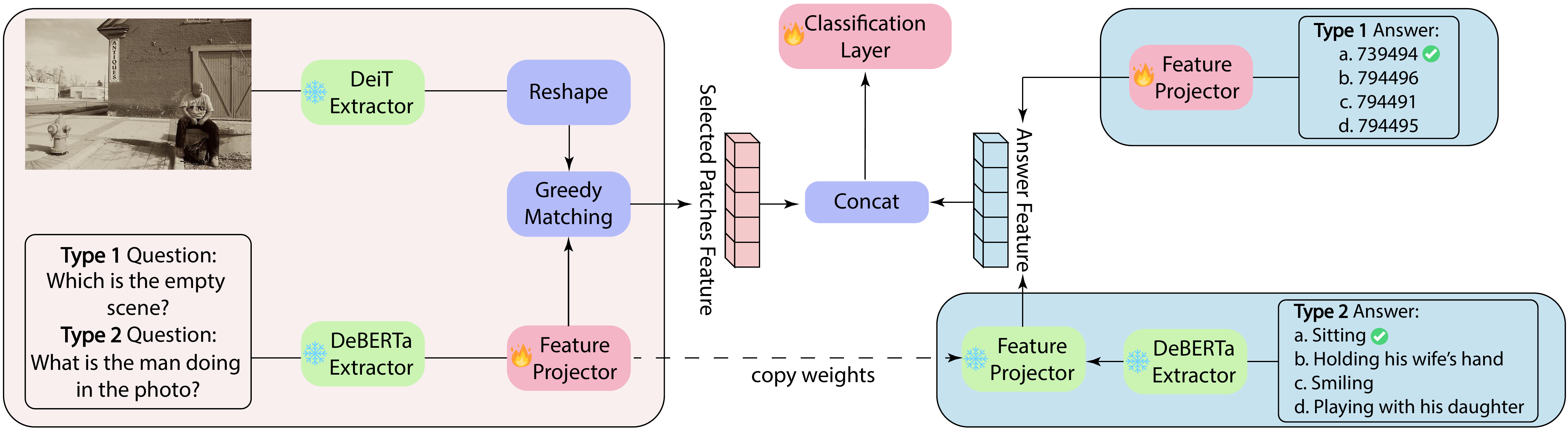}
\end{center}
\caption{Overview of \texttt{ProtoVQA}. \texttt{ProtoVQA} extracts visual features through DeiT Extractor and encodes questions through DeBERTa Extractor. Image patch features undergo greedy matching with question-aware prototypes generated through a feature projector. For answering, the model processes either coordinate inputs (Type 1) through another projector, or textual answers (Type 2) through DeBERTa and a frozen feature projector sharing weights from the question branch. The matched patch features are concatenated with answer features for final classification.}
\label{fig:model}
\end{figure*}

Prototype-based learning has emerged as a promising approach to improving interpretability \cite{chen2019looks, barnett2021case, donnelly2022deformable, ma2023looks}. The latest work like ProtoViT \cite{ma2024interpretable} shows that Vision Transformers can enable flexible prototype learning while maintaining interpretability. In ProtoViT, each prototype is a learned embedding representing a recurring semantic concept, such as an object part or texture. At inference, the model matches image patches to prototypes and visualizes these regions to expose the reasoning process. For example, a “beak” prototype often activates on bird-head areas, grounding the decision in interpretable visual cues.

While these methods have shown success in purely visual tasks, extending prototype-based reasoning to multimodal settings introduces unique challenges. In particular, VQA requires aligning visual evidence with language queries, which complicates prototype learning and interpretability. Key challenges include:
\textit{(i)} Prototype-based approaches often focus on single modality (visual/textual) interpretability, struggling to bridge the visual-textual semantic gap;
\textit{(ii)} Rigid prototypical features fail to capture geometric variations and dynamic visual-question relationships;
\textit{(iii)} These methods lack the ability to provide fine-grained explanations at both the component level and system level, making it difficult to understand how individual parts contribute to the final decision.

To address these issues, we propose \texttt{ProtoVQA}. Our contributions are:
\begin{itemize}[leftmargin=*]

\item We introduce an adaptable prototypical framework capable of seamlessly handling diverse visual-linguistic downstream tasks, including both visual question answering and grounding, through a shared prototype-based backbone with task-specific answer processing modules.

\item We employ a spatially-constrained greedy matching strategy to model dynamic visual-question relationships and geometric variations.

\item Our model achieves comprehensive explainability through explicit visual evidence and systematic validation of visual-linguistic alignment.
 
\end{itemize}

\section{\texttt{ProtoVQA}}
We present \texttt{ProtoVQA} (Figure \ref{fig:model}), a prototype-based approach to visual question answering that achieves interpretability through question-aware prototype learning and spatially-constrained patch matching. By explicitly mapping prototypes to discriminative image regions, \texttt{ProtoVQA} can provide transparent reasoning paths from questions to visual evidence.

\subsection{Feature Extraction Module}
The visual feature extraction leverages pre-trained DeiT~\cite{touvron2021training} as a backbone to extract patch-level visual features. Let $I \in \mathbb{R}^{H \times W \times 3}$ denote the input image. The DeiT backbone processes $I$ to produce a feature map $F = [f_{\mathrm{CLS}}, f_1, \ldots, f_N] \in \mathbb{R}^{(N+1) \times D}$, where $f_{\mathrm{CLS}} \in \mathbb{R}^D$ is the global CLS token feature and $f_n \in \mathbb{R}^D$ for $n \in \{1,...,N\}$ are the image patch features. Patch feature representation is enhanced by forming $F_{\mathrm{visual}} = [f_1 - f_{\mathrm{CLS}},\ \ldots,\ f_N - f_{\mathrm{CLS}}]$.

For textual input, \texttt{ProtoVQA} utilizes a pre-trained DeBERTa model~\cite{he2021deberta}. The question $Q$, represented as a token sequence $[q_1, q_2, ..., q_{l_q}]$, is encoded by DeBERTa, yielding embeddings $E_q \in \mathbb{R}^{l_q \times D_{\text{text}}}$, where $D_{\text{text}}$ is the DeBERTa hidden dimension. These embeddings are then projected into the shared visual-linguistic space $\mathbb{R}^D$ via a learnable feature projector $\mathcal{F}$. For answer processing, there are two pathways: For question answering tasks, answer candidates are encoded by DeBERTa and projected to $\mathbb{R}^D$ using the same feature projector $\mathcal{F}$ with frozen parameters whose weights are copied from the question encoding. This weight-sharing mechanism ensures consistent representation of question and answer candidates in the shared visual-linguistic space, while the frozen parameter design prevents potential overfitting. For visual grounding tasks, the coordinate inputs $P \in \mathbb{R}^4$ are directly projected to the same feature space through a separate feature projector.

\subsection{Interpretable Prototypical Part Selection Module}
This module constitutes the core novelty and interpretability mechanism of \texttt{ProtoVQA}. It introduces sub-patch prototypes and a greedy matching algorithm with spatial constraints to select salient image parts.

\subsubsection{Sub-patch Prototypes}
To link question semantics with visual evidence, we reshape the first $m \times k$ projected question tokens into a 3D tensor:
\begin{equation}
P = \text{Reshape}(\mathcal{F}(E_q[:m \times k])) \in \mathbb{R}^{m \times k \times D}.
\end{equation}
This forms $m$ prototypes, each composed of $k$ sub-patches prototypes sharing the same dimensionality as visual features. These prototypes act as semantic anchors, capturing discriminative visual concepts conditioned on the question. A learnable weighting mechanism modulates the relevance of each sub-patch, enabling context-aware patch selection. The resulting prototypes are then used in the matching process (Section \ref{greedy_selection}), where their alignment with the image regions directly influences the final prediction.

\subsubsection{Greedy Matching for Patch Selection}
\label{greedy_selection}
The core matching mechanism employs a spatially-constrained greedy algorithm \cite{ma2024interpretable} to establish correspondences between sub-patch prototypes and image regions. For each prototype $P_i \in \mathbb{R}^{k \times D}$ from our prototype set $P$, the algorithm iteratively constructs a spatially coherent set of matched image patches through $k$ iterations.

At each iteration $t$, we first calculate the similarity matrix $S^t \in \mathbb{R}^{N \times k}$ between all patch features $F_{\mathrm{visual}}$ and prototype sub-patches $P_i$:

\begin{equation}
    S^t_{n,j} = \frac{F_{\mathrm{visual},n} \cdot P_{i,j}}{\|F_{\mathrm{visual},n}\| \|P_{i,j}\|},
\end{equation}
where $n \in \{1,...,N\}$ indexes image patches and $j \in \{1,...,k\}$ indexes sub-patches.

The algorithm then identifies the optimal patch-subpatch pair $(n^*, j^*)$ that maximizes the similarity score:
\begin{equation}
    (n^*, j^*) = \operatorname*{argmax}_{n,j} \; S^t_{n,j} \cdot M^t_{n} \cdot A^t_{n},
\end{equation}
where $M^t \in \{0,1\}^N$ is a binary mask that indicates available patches in iteration $t$ ($1$ for available patches, $0$ for unavailable), and $A^t \in \{0,1\}^N$ is an adjacency mask that enforces spatial continuity with previously selected patches. After each selection, the masks are updated: $M^{t+1}$ marks the selected patch as unavailable by setting $M^{t+1}_{n^*} = 0$ to prevent repeated selection in subsequent iterations, and $A^{t+1}$ is updated to mark as valid only those patches within a spatial constraint radius $r$ from position $n^*$, ensuring spatial coherence in the matching process.

The final matching score for prototype $P_i$ is computed as a weighted combination of individual sub-patch similarities:
\begin{equation}
    \text{score}(P_i) = \sum_{t=1}^k w_t \cdot S^t_{n^*_t,j^*_t},
\end{equation}
where $w_t$ are learnable slot weights that modulate the importance of each sub-patch match, and $(n^*_t,j^*_t)$ denotes the optimal pair selected at iteration $t$. This spatially-aware matching strategy ensures the selected patches form coherent visual regions while maintaining semantic relevance to the prototype.

\subsection{Answer Processing}
\texttt{ProtoVQA} supports two types of answer processing: \textbf{Type 1 (Visual Grounding)} for tasks requiring precise coordinate-based answers, where input coordinates $P \in \mathbb{R}^4$ are projected directly into the feature space through a dedicated projector; and \textbf{Type 2 (Descriptive QA)} for tasks requiring textual answers, where candidates are encoded by DeBERTa and projected using a frozen feature projector that shares weights with the question branch, ensuring consistent representation while preventing overfitting. In both cases, the matched patch features are concatenated with the processed answer features and fed directly through a classification layer for final prediction.

\begin{table*}[t]
\centering
\scalebox{0.9}{
\begin{tabular}{c!{\vrule width \lightrulewidth}ccc} 
\toprule
Method          & Vision Encoder & Text Encoder & Accuracy (\%)$\uparrow$  \\ 
\midrule
SUPER  \cite{han2023semantic}                  & FasterRCNN             & GRU                   & 64.07              \\
QOI\_Attention  \cite{gao2018examine}         & FasterRCNN             & GRU                   & 65.90              \\
SDF of VLT  \cite{ding2022vlt}             & ViT-patch16            & BERT                  & 65.93              \\
STL   \cite{wang2018structured}                   & ResNet200              & n-gram                & 68.20              \\
CFR   \cite{nguyen2022coarse}                   & FasterRCNN             & GRU                   & 71.90               \\
BriVL     \cite{fei2022towards}               & Custom image patch+CNN & RoBERTa               & 72.06              \\
CTI        \cite{do2019compact}              & FasterRCNN             & LSTM/GRU              & 72.30               \\
Bi-CMA \cite{upadhyay2025bidirectional}                & ViT-patch16            & BERT                  & 70.53              \\
Bi-CMA  \cite{upadhyay2025bidirectional}            & ViT-patch16 (finetune) & BERT                  & 73.07              \\ 
\midrule
\textbf{\texttt{ProtoVQA} (Ours)} & ViT-patch16            & DeBERTa               & 70.23              \\
\bottomrule
\end{tabular}
}
\caption{
Accuracy comparison with representative state-of-the-art foundation VQA models on the Visual7W \cite{zhu2016visual7w} test set. The table lists the vision encoder backbone, the text encoder backbone, and the accuracy of each method, with accuracy reported in percentages (indicated by the \% symbol). The evaluated methods span both traditional CNN–RNN pipelines and modern Transformer-based vision–language architectures. The result of \texttt{ProtoVQA} is shown in the final row. The $\uparrow$ symbol indicates that higher values represent better performance.
}
\label{tab:Visual7W_result}
\end{table*}

\subsection{Visual–Linguistic Alignment Evaluation}
A key requirement for interpretable VQA is that models not only produce the correct answer but also ground their reasoning process in semantically relevant visual evidence. 
Existing evaluation protocols typically rely on Intersection-over-Union (IoU) or other pixel-level overlap metrics that are primarily designed for detection or segmentation tasks. 
While effective for measuring localization accuracy, these metrics are poorly aligned with the goal of explanation because they may penalize valid but partially overlapping evidence, are highly sensitive to annotation scale, and fail to capture whether the selected regions are conceptually meaningful in the context of the question. 

To address these limitations and systematically evaluate the alignment between visual and linguistic components, we introduce the \textbf{Visual–Linguistic Alignment Score (VLAS)}, an interpretability-oriented metric that directly evaluates whether a model’s attended regions are semantically consistent with the ground-truth evidence. 
For each QA pair $i$, we compute:
\begin{equation}
\text{VLAS} = \frac{\sum_{i=1}^{N} \mathcal{I}(M_i \cap G_i > \theta)}{N_{QA}},
\end{equation}
where $M_i$ denotes the model-attended region, represented as the union of matched patch boxes, and $G_i$ is the corresponding ground truth region. The indicator function $\mathcal{I}(\cdot)$ returns 1 if the intersection-over-union (IoU) between $M_i$ and $G_i$ exceeds the threshold $\theta$, and 0 otherwise. We follow standard practice in object detection and set $\theta = 0.5$.

VLAS offers several advantages compared with traditional IoU-based metrics: 
(i) it captures the binary nature of human judgments by measuring whether an explanation is acceptable rather than rewarding incremental overlap; 
(ii) it mitigates biases caused by variable annotation scales; 
and (iii) it enables robust dataset-level evaluation by aggregating the proportion of QA pairs with satisfactory alignment instead of averaging raw IoU values, which can be skewed by a few large regions.

\begin{figure*}[htbp]
    \centering
    \subfigure[Which item can be used for communication?]{
        \includegraphics[width=0.233\textwidth]{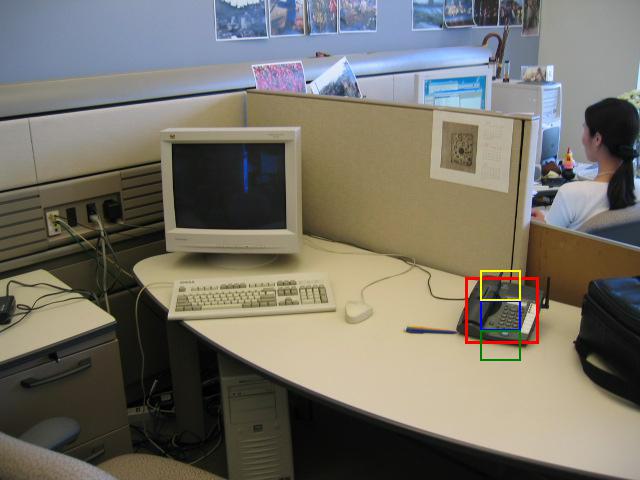}
        \label{fig:2a}
    }
    \subfigure[Which is framing a white sideways boat?]{
        \includegraphics[width=0.233\textwidth]{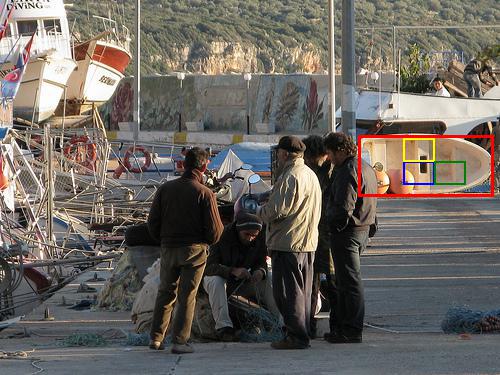}
        \label{fig:2b}
    }
    \subfigure[Which ear is the left ear of the right giraffe?]{
        \includegraphics[width=0.233\textwidth]{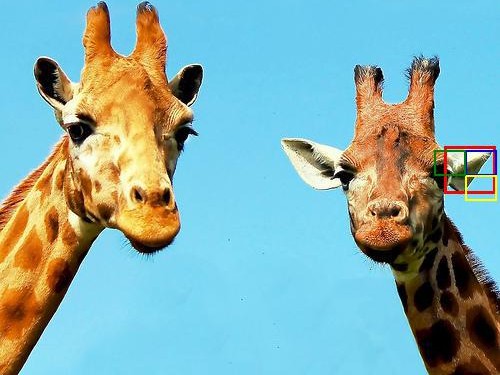}
        \label{fig:2c}
    }
    \subfigure[Which flower tub, with red flowers in it, is beside a parking meter?]{
        \includegraphics[width=0.233\textwidth]{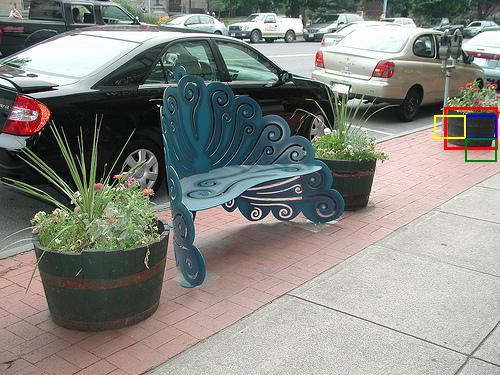}
        \label{fig:2d}
    }
    \caption{Visualization of explanation results on Visual7W \cite{zhu2016visual7w} test set. The \textcolor{red}{\textbf{red}} bounding box indicates the ground truth answer box provided by the dataset. The \textcolor{blue}{\textbf{blue}}, \textcolor{dartgreen}{\textbf{green}} and \textcolor{yellow}{\textbf{yellow}} bounding boxes show the \textbf{top-3 best-matched patches} projected back to the original image space. More visualization results on diverse visual question answering scenarios can be found in Appendix Section \ref{additional_visual}.}
    \label{fig:visual_results}
\end{figure*}

\section{Experiments}
\label{exp}

\subsection{Setup}

\paragraph{Dataset}
\underline{\textit{(i)}} Visual7W~\cite{zhu2016visual7w} is a large-scale grounded VQA benchmark comprising 327{,}939 question–answer pairs collected over 47{,}300 COCO images. 
Each question is paired with four human-curated multiple-choice options, totaling over 1.3M candidate answers, and 561{,}459 object-level groundings spanning 36{,}579 unique categories. 
These rich annotations enable fine-grained evaluation of both answer prediction and visual grounding, making Visual7W a widely adopted and challenging benchmark for studying multimodal reasoning.

\paragraph{Baselines} Detailed descriptions for all baselines are provided in Appendix Section~\ref{baseline}.

\paragraph{Configuration}
The model was trained on an NVIDIA A800 GPU (80GB) for 200 epochs using Adam optimizer (lr=$1\times10^{-4}$, batch size=64). The vision transformer processed $224\times224$ images with $16\times16$ patches. The prototype learning used $m=10$ prototypes per class (each with $k=3$ sub-patches prototypes) and a spatial constraint radius of $r=3$. Other hyperparameters remained default.

\subsection{Comparison with Baselines}
As shown in Table~\ref{tab:Visual7W_result}, among the methods using ViT-patch16 as the visual backbone, \texttt{ProtoVQA} (70.23\%) achieves performance comparable to Bi-CMA (70.53\% without fine-tuning, 73.07\% with fine-tuning) and outperforms the SDF of VLT (65.93\%). This demonstrates that our framework, while primarily designed to provide transparent and interpretable reasoning, still delivers accuracy that remains within the expected range of strong Transformer-based baselines. In particular, although fine-tuned Bi-CMA achieves slightly higher performance, \texttt{ProtoVQA} offers the additional advantage of prototype-grounded explanations, showing that interpretability can be introduced without incurring a substantial drop in competitiveness.

\subsection{Qualitative Visualization}
Figure~\ref{fig:visual_results} provides qualitative examples from the Visual7W test set, showing how \texttt{ProtoVQA} grounds its reasoning in semantically relevant image regions. The model-attended regions (\textcolor{blue}{\textbf{blue}}, \textcolor{dartgreen}{\textbf{green}}, and \textcolor{yellow}{\textbf{yellow}} boxes) align closely with the dataset-provided ground-truth annotations (\textcolor{red}{\textbf{red}} boxes).

For instance, in Figure~\ref{fig:2a}, when asked which item can be used for communication, the model correctly highlights the telephone region, focusing on the same area as the ground truth. In Figure~\ref{fig:2b}, the model identifies the frame of the sideways boat, with matched patches overlapping the annotated boundary. In Figure~\ref{fig:2c}, for the question about the giraffe’s left ear, the model consistently selects patches concentrated on the ear region, demonstrating fine-grained part-level reasoning. In Figure~\ref{fig:2d}, the model highlights the flower tub with red flowers near the parking meter, showing its ability to handle relational queries that involve both object attributes and spatial context.

Overall, these qualitative examples demonstrate that \texttt{ProtoVQA} consistently grounds its answers in semantically relevant visual evidence across diverse scenarios, thereby providing explanations that are both human-verifiable and closely aligned with the questions.

\subsection{Evaluation of Visual–Linguistic Alignment}

\begin{table}[htbp]
\centering
\scalebox{0.8}{
\begin{tabular}{ccc}
\toprule
Method  & VLAS@1$\uparrow$  & VLAS@3$\uparrow$  \\
\midrule
SDF of VLT   &    0.2013     &    0.0847    \\
Bi-CMA   &     0.2466    &   0.1123     \\
\textbf{\texttt{ProtoVQA} (Ours)}    &  0.4103      &   0.2466     \\
\bottomrule
\end{tabular}
}
\caption{
Visual–Linguistic Alignment Score (VLAS) on Visual7W \cite{zhu2016visual7w}. 
Values for VLAS@1 and VLAS@3 are reported, with $\uparrow$ indicating that higher scores correspond to better alignment. 
}
\vspace{-0.3cm}
\label{vlas_comparison}
\end{table}

As shown in Table \ref{vlas_comparison}, \texttt{ProtoVQA} significantly outperforms baseline methods on VLAS (0.4103 vs 0.2466 on VLAS@1, 0.2466 vs 0.1123 on VLAS@3), representing a 66.4\% and 119.6\% improvement over Bi-CMA respectively, thereby clearly and consistently demonstrating superior visual-linguistic alignment capability.

\section{Conclusion}
We present \texttt{ProtoVQA}, a novel framework for visual question answering that addresses the need for model transparency and cross-modal reasoning. \texttt{ProtoVQA} achieves comprehensive explainability by (i) introducing adaptable prototypes capable of seamlessly handling diverse visual-linguistic downstream tasks through a shared prototype-based backbone; (ii) employing a spatially-constrained greedy matching strategy to model dynamic visual-question relationships and geometric variations; and (iii) providing explicit visual evidence and systematic validation of visual-linguistic alignment.  Our work provides a fundamental step towards VQA systems that achieve strong performance while maintaining comprehensive explainability.

\section*{Ethical Considerations}
We examined the study describing the publicly available datasets used in this research and identified no ethical issues regarding the datasets. 

\section*{Acknowledgment}
This study is supported by the Department of Defense under Grant No. HT9425-23-1-0267 and in part by the National Science Foundation under Grant No. 2452367.

\clearpage
\newpage

\section*{Limitations}

While this study shows promising results, several limitations remain. (1) Although \texttt{ProtoVQA} provides comprehensive interpretability, improving the faithfulness of prototype-based explanations under the constraint of preserving task performance remains an open problem. Future work could explore jointly optimized training objectives, adaptive prototype initialization, or more expressive matching strategies to balance accuracy and transparency. (2) Our evaluation is restricted to general-purpose VQA benchmarks; transferring the framework to domain-specific or safety-critical settings (e.g., medical imaging, autonomous driving) may require curating specialized prototype vocabularies, domain-adaptive calibration, and additional fine-tuning to account for distributional shifts. (3) The current architecture is designed for multiple-choice and grounding-style tasks and has not yet been extended to prompt-based or free-form generative VQA supported by large language models. Integrating prototype reasoning with instruction-tuned generative models and multi-step reasoning pipelines is a promising direction for enabling more general and scalable interpretability. We leave these challenges for future work, aiming to advance VQA systems that achieve strong performance while offering faithful and transparent reasoning.

\bibliography{ref}
\appendix

\section{Baselines}
\label{baseline}
\begin{itemize}[leftmargin=*]
    \item {
    \textbf{SUPER} \cite{han2023semantic}: Introduces a semantic-aware modular capsule routing framework for Visual Question Answering (VQA) to enhance adaptability to semantically complex inputs. It features five specialized modules and dynamic routers that refine vision-semantic representations, offering a novel approach to architecture learning and representation calibration for VQA tasks.
    }
    \item {
    \textbf{QOI\_Attention} \cite{gao2018examine}: Proposes a Multi-task Learning with Adaptive-attention (MTA) model for multiple-choice (MC) VQA. It mimics human reasoning by integrating answer options and adapting attention to visual features, achieving remarkable performance on MC VQA benchmarks.
    }
    \item {
    \textbf{SDF of VLT} \cite{ding2022vlt}: Presents a Vision-Language Transformer (VLT) framework for referring segmentation, introducing a Query Generation Module to dynamically produce input-specific queries. It improves handling diverse language expressions with a Query Balance Module and masked contrastive learning, setting new benchmarks on five datasets.
    }
    \item {
    \textbf{STL} \cite{wang2018structured}: Proposes a VQA model focused on the multiple-choice task, incorporating part-of-speech (POS) tag-guided attention, convolutional n-grams, and triplet attention interactions between the image, question, and candidate answer. The model also employs structured learning for triplets based on image-question pairs.
    }
    \item {
    \textbf{CFR} \cite{nguyen2022coarse}: Introduces a reasoning framework for Visual Question Answering (VQA) that bridges the semantic gap between image and question by jointly learning features and predicates in a coarse-to-fine manner. The model achieves superior VQA accuracy and provides an explainable decision-making process.
    }
    \item {
    \textbf{BriVL} \cite{fei2022towards}: Develops a foundation model pre-trained on multimodal data for artificial general intelligence (AGI), focusing on self-supervised learning with weak semantic correlation data. The model demonstrates strong imagination ability, achieving promising results across various downstream tasks including VQA.
    }
    \item {
    \textbf{CTI}~\cite{do2019compact}: Introduces a trilinear interaction model for Visual Question Answering (VQA) to learn associations between image, question, and answer modalities, using PARALIND tensor decomposition for efficiency. For free-form VQA, knowledge distillation transfers learnings to a bilinear student model, achieving state-of-the-art results on TDIUC and Visual7W datasets.
    }
    \item {
    \textbf{Bi-CMA} \cite{upadhyay2025bidirectional}: Proposes a Bidirectional Cascaded Multimodal Attention network for VQA, utilizing bidirectional attention and sparsity to enhance feature integration between image and text. The model performs competitively on multiple-choice VQA tasks, providing insightful attention maps that reveal the model's decision-making focus.
    }
\end{itemize}

\section{Greedy Matching}
We adopted greedy matching as a simple yet efficient baseline to demonstrate how our framework functions in practice. The strategy selects the region with the highest similarity to a prototype at each step, making it computationally inexpensive, easy to implement, and naturally interpretable since every decision can be visualized as part of the reasoning trail. Its low complexity also makes it suitable for large-scale experiments and real-time inference, where maintaining both speed and transparency is crucial. This balance of efficiency, scalability, and interpretability highlights why greedy matching serves as a strong and practical choice for validating the effectiveness of our prototype-based framework. In addition, it establishes a clear benchmark that future improvements can be directly compared against. This makes greedy matching an integral component in demonstrating the overall practicality of our approach.

\section{Potential Applications}
Interpretable VQA has broad potential across domains where both accurate answers and transparent reasoning are essential. In media forensics, explanation-aware VQA can help detect and verify manipulated or misleading short videos by aligning visual evidence with textual claims \cite{wang2025consistency, wang2025fakesv, gao2024gem}. In transportation and civil engineering, interpretable models can support safety-critical decisions, such as predicting pavement conditions from visual cues while providing human-verifiable justifications \cite{lu2025journey,lu2025predicting}. For general machine learning tasks, interpretable VQA can benefit incomplete or multi-view multi-label classification \cite{xie2024uncertainty,xie2025multi} as well as multimodal representation learning, where pruning, efficient adaptation, and alignment with language supervision remain active directions \cite{guo2025crop, liu2024clips, jian2023bootstrapping, liu2023mllms, chen2024taskclip,liu2025modality,zhang2025pretrained}. In video understanding, explanation-guided alignment can improve temporal grounding, multimodal reasoning, and working memory in instructional or complex video scenarios \cite{li2024towards, diao2025temporal, zhou2025glimpse}. In human–AI interaction and robotics, transparent reasoning is crucial for building trustworthy assistants that combine long-horizon planning and personalized interaction with multimodal evidence \cite{zhang2025overcoming, zhou2025language, xiang2025regrap}. Finally, in creative and cultural applications, interpretable VQA can support tasks such as music performance understanding and question answering, or semantic analysis of non-standard scripts and pictograms, where human-verifiable explanations are indispensable for reliability and adoption \cite{you2025music, diao2025learning, bi2025dongbamie}.

\section{Additional Visualization Results}
\label{additional_visual}
In addition to the results shown in Figure \ref{fig:visual_results} in Section \ref{exp}, we further provide 10 representative samples from the Visual7W \cite{zhu2016visual7w} test set to illustrate the breadth of interpretability achieved by \texttt{ProtoVQA}. These additional cases cover a wide spectrum of question types and visual reasoning demands, offering a more complete view of how the model grounds its predictions in semantically meaningful evidence. 

\begin{figure*}[ht]
\begin{center}
\includegraphics[width=1\linewidth]{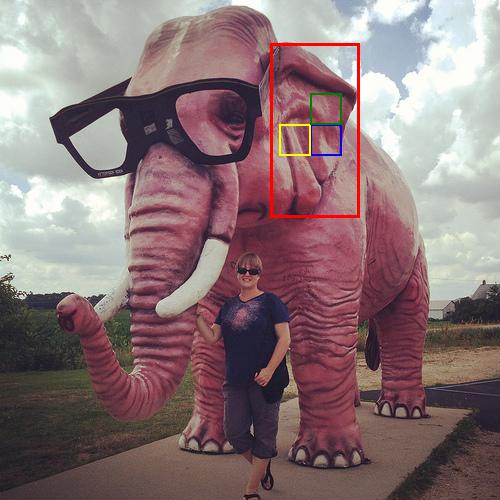}
\end{center}
\caption{Question: Which part helps the elephant hear?}
\label{fig:case10}
\end{figure*}

\begin{figure*}[t]
\begin{center}
\includegraphics[width=0.9\linewidth]{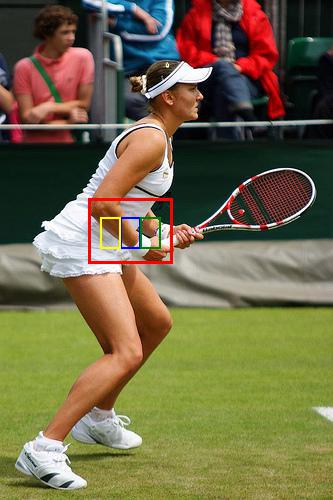}
\end{center}
\caption{Question: Which is the players arms?}
\label{fig:case1}
\end{figure*}

\begin{figure*}[ht]
\begin{center}
\includegraphics[width=1\linewidth]{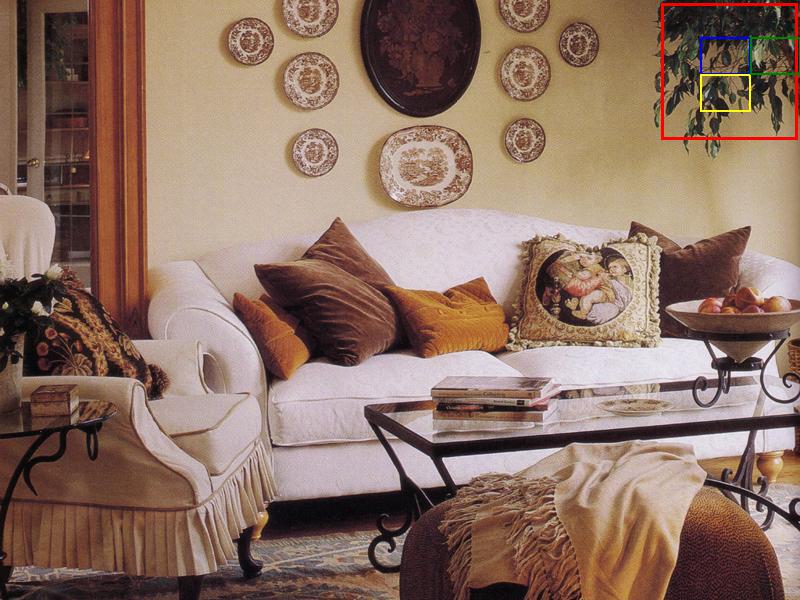}
\end{center}
\caption{Question: Which plant is hanging in the room?}
\label{fig:case2}
\end{figure*}

\begin{figure*}[ht]
\begin{center}
\includegraphics[width=0.9\linewidth]{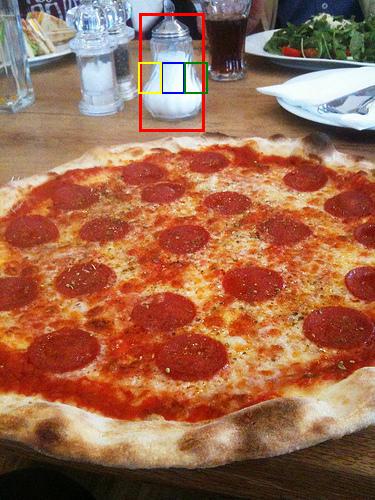}
\end{center}
\caption{Question: Which is the glass containing?}
\label{fig:case5}
\end{figure*}

\begin{figure*}[ht]
\begin{center}
\includegraphics[width=1\linewidth]{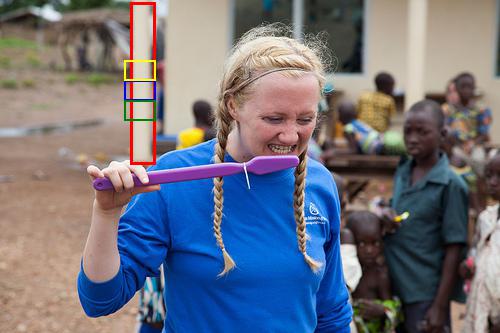}
\end{center}
\caption{Question: Which object is a large beige cylinder next to the dirt?}
\label{fig:case7}
\end{figure*}

\begin{figure*}[ht]
\begin{center}
\includegraphics[width=1\linewidth]{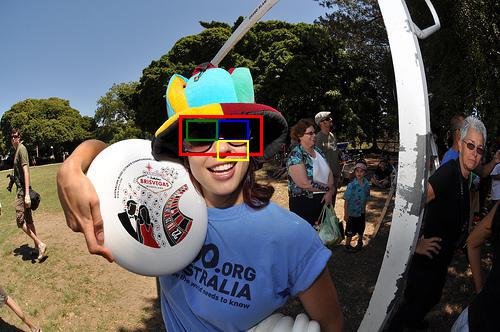}
\end{center}
\caption{Question: Which object is she wearing on her face?}
\label{fig:case3}
\end{figure*}

\begin{figure*}[ht]
\begin{center}
\includegraphics[width=0.8\linewidth]{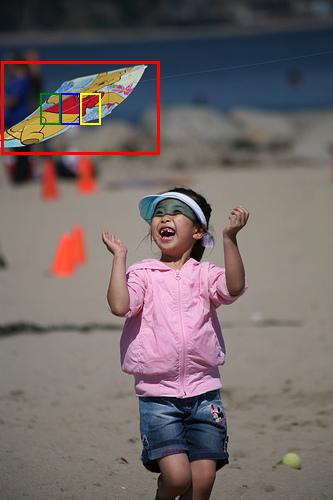}
\end{center}
\caption{Question: Which object is being flown?}
\label{fig:case6}
\end{figure*}

\begin{figure*}[ht]
\begin{center}
\includegraphics[width=1\linewidth]{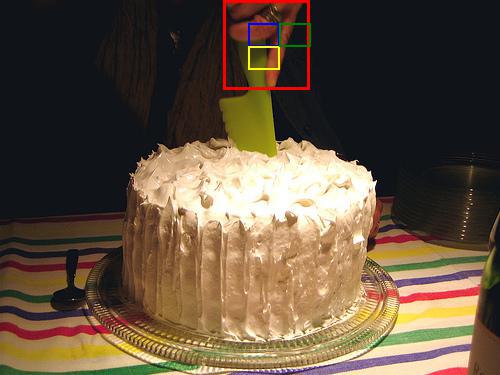}
\end{center}
\caption{Question: Which hand is holding a knife?}
\label{fig:case8}
\end{figure*}

\begin{figure*}[ht]
\begin{center}
\includegraphics[width=1\linewidth]{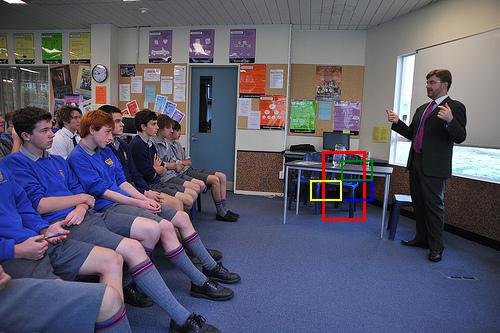}
\end{center}
\caption{Question: Which blue chair behind the table?}
\label{fig:case4}
\end{figure*}

\begin{figure*}[ht]
\begin{center}
\includegraphics[width=1\linewidth]{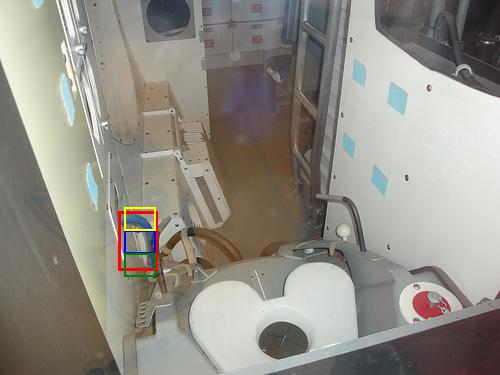}
\end{center}
\caption{Question:  Which hose is sticking out of the wall?}
\label{fig:case9}
\end{figure*}

Specifically, the examples span multiple categories of visual–linguistic reasoning. For questions involving human and animal anatomy (Figures \ref{fig:case10}, \ref{fig:case1}), the model is able to precisely highlight fine-grained parts, such as arms or ears, showing that prototype matching is sensitive to localized semantic cues. For object identification tasks (Figures \ref{fig:case2}, \ref{fig:case5}, \ref{fig:case7}), the model consistently selects patches that coincide with the relevant object boundaries, even when distractors are present in the scene. For interaction-related questions (Figures \ref{fig:case3}, \ref{fig:case6}, \ref{fig:case8}), the highlighted regions demonstrate the model’s ability to capture contextual relationships, such as a person holding an item or an object being actively manipulated. Finally, in spatial relationship queries (Figures \ref{fig:case4}, \ref{fig:case9}), the attended patches illustrate how the model disambiguates relative positions, grounding its answer in spatially coherent regions.

Overall, these qualitative examples highlight that \texttt{ProtoVQA} is not limited to generic visual cues but adapts its evidence selection to the specific semantics of each question. The consistency between the model-attended patches and the dataset-provided ground truth shows that the framework provides reliable, human-verifiable explanations across a diverse set of VQA scenarios, further validating the interpretability and robustness of our approach.

\end{document}